\pdfoutput=1

\documentclass[11pt]{article}

\usepackage{ACL2023}

\usepackage{times}
\usepackage{latexsym}
\usepackage{graphicx}

\usepackage[T1]{fontenc}

\usepackage[utf8]{inputenc}

\usepackage{microtype}

\usepackage{inconsolata}

\usepackage{booktabs}
\usepackage{lipsum}  
\usepackage{makecell}
\usepackage{array}
\newcolumntype{?}{!{\vrule width 1.2pt}}
\usepackage{mathtools}
\usepackage{siunitx}
\usepackage{multirow}
\usepackage{enumitem}
\usepackage{hhline}

\usepackage{caption}
\usepackage{multicol}
\newcommand*\RETRO{\textsc{Retro}}
\newcommand*\RETROON{\textsc{Retro[on]}}
\newcommand*\RETROOFF{\textsc{Retro[off]}}
\newcommand*{\BMXXV}{\texttt{BM25}}
\newcommand*{\BPB}{\textsc{Bpb}}
\newcommand*{\PPL}{\textsc{Ppl}}

\title{Surface-Based Retrieval Reduces Perplexity of\\Retrieval-Augmented Language Models}

\def\authorsep{\hspace{0.3em}}

\author{Ehsan Doostmohammadi$^{1}$\thanks{\ \ Correspondence to \href{mailto:ehsan.doostmohammadi@liu.se}{\texttt{ehsan.doostmohammadi@liu.se}}.} \authorsep Tobias Norlund$^{2,4}$ \authorsep Marco Kuhlmann$^{1}$ \authorsep Richard Johansson$^{2,3}$\\
\null$^{1}$ Linköping University \quad \null$^{2}$ Chalmers University of Technology \\
\null$^{3}$ University of Gothenburg \quad \null$^{4}$ Recorded Future}

\begin{document}
\maketitle

\begin{abstract}
Augmenting language models with a retrieval mechanism has been shown to significantly improve their performance while keeping the number of parameters low.
Retrieval-augmented models commonly rely on a semantic retrieval mechanism based on the similarity between dense representations of the query chunk and potential neighbors.
In this paper, we study the state-of-the-art \RETRO\ model and observe that its performance gain is better explained by surface-level similarities, such as token overlap.
Inspired by this, we replace the semantic retrieval in \RETRO\ with a surface-level method based on \BMXXV, obtaining a significant reduction in perplexity.
As full \BMXXV\ retrieval can be computationally costly for large datasets, we also apply it in a re-ranking scenario, gaining part of the perplexity reduction with minimal computational overhead.
\end{abstract}

\section{Introduction}

The introduction of the Transformer architecture \citep{transformer} has led to a performance boost in language modeling (see, e.g., \citealt{brown2020language}), but also to a steep increase of computational cost, as the number of parameters and data points is constantly growing.
In reaction to this development, there has recently been a surge in work on retrieval-augmented language models \cite{izacard-grave-2021-leveraging, li2022survey}, which shows that enabling models to retrieve context from large corpora results in lower perplexity and better accuracy in downstream tasks such as question answering, while at the same time using considerably fewer parameters.
In this paper, we specifically focus on the Retrieval-Enhanced Transformer architecture (\RETRO; \citealp{retro}).

By augmenting a language model with a retrieval mechanism, \RETRO, like similar architectures, tries to decouple \emph{memorization} of the training data from the additional \emph{generalization} that comes with increasing the number of parameters.
In \RETRO, when a chunk of text (a sequence of tokens) has been generated, a dense representation of this chunk is used to retrieve the most similar neighboring chunks from a large retrieval set, based on their L2 distance.
Having the previously generated chunks and their nearest neighbors in the retrieval set, the auto-regressive language model has now access to an extended context when predicting the next chunk.
The informativeness of this context depends on the effectiveness of the retrieval method.

\citet{retro} note that part of \RETRO's performance can be attributed to the token overlap between the generated chunks and the retrieval set.
Our starting point in this paper is the observation that the performance gain is actually \emph{better} explained by such surface-level similarities than by the L2 distance between the dense representations that \RETRO\ uses for retrieval.
This is in line with recent work by \citet{norlund-etal-2023-generalization}, who show that the reduction in loss observed in \RETRO\ ``almost exclusively'' stems from such overlap rather than more sophisticated generalization.
Based on these findings, we replace the semantic retrieval method in \RETRO\ with one based on \BMXXV\ \cite{robertson1995okapi}, a surface-level measure.
Our results show that retrieving nearest neighbors using \BMXXV\ during inference leads to a 13.6\% lower perplexity, compared to dense retrieval based on sentence transformers (ST) \cite{reimers-2019-sentence-bert}, a model trained to represent the semantic similarity between sentences.\footnote{The code and the data for this study can be accessed at \href{https://github.com/edoost/retro_bm25}{\texttt{github.com/edoost/retro\_bm25}}.}

Finding the exact neighbors with \BMXXV\ is costly on large retrieval sets and might not meet the speed requirements of all applications of retrieval-augmented language models.
We therefore explore a hybrid approach where we first retrieve approximate neighbors using ST representations and then re-rank them using \BMXXV.
We show that this approach yields 24.7\% of the perplexity reduction we get with \BMXXV-based retrieval, with only minimal computational overhead.



\section{Method}

We experiment with \RETRO\ \citep{retro} as a state-of-the-art retrieval-augmented language model.

\subsection{Model}

\RETRO\ is very similar to a standard auto-regressive language model such as T5 \cite{raffel2020t5}, the main differences being the introduction of the retrieval mechanism and how the retrieved neighbors are used for language modeling.

\paragraph{Nearest Neighbor Retrieval}

In \RETRO, all textual data is stored and used in chunks of 64 tokens. When the model has generated a chunk $C_u$, it retrieves the $k$ nearest neighbors $N_{1:k}$ to that chunk, together with the chunks $F_{1:k}$ following these neighbor chunks in the retrieval data. It then generates the next chunk $C_{u+1}$ conditioned on the retrieved chunk pairs. Retrieval uses the squared L2 distance on a dense representation ($\mathit{DR}$) of chunks:
\[
 d(C_u, N_i) = \|\mathit{DR}(C_u) - \mathit{DR}(N_i)\|^2_2
\]
This leaves us with
\[
\textsc{Ret}(C_u) = ([N_u^1; F_u^1], \dots, [N_u^k; F_u^k])
\]
as the retrieved neighbors that the model receives as additional context when generating the next chunk. The likelihood of the first chunk ($C_1$) does not depend on any neighbors; the model has access to no external context when generating that chunk. During training and perplexity evaluation, the retrieval process is filtered such that chunks originating from the same source document as the training sequence are never considered as neighbors.

\paragraph{Integration of the Neighbors}

\RETRO\ improves auto-regressive language modeling by conditioning the next token prediction on the retrieved chunks of text.
This means that the probability of generating the next token $x_{t+1}$ depends not only on the previously generated tokens $x_{1:t}$ but also on the retrieved neighbors of the previously generated chunks, as well as their following chunks:
\[
P\left(x_{t+1} \,|\, x_{1:t}, \textsc{Ret}(C_1), \dots, \textsc{Ret}(C_{u-1}); \theta \right)
\]

When generating the next token, the neighbors as well as the current chunk $C_u$ are passed through a Transformer encoder.
In the decoder, cross-attention is over the output of that encoder and the concatenation of the intermediary embeddings of the last few tokens in the previous chunk $C_{u-1}$ and the already generated tokens in $C_u$, a mechanism called \emph{chunked cross-attention}.
For more details, see \citet{retro}.

\paragraph{Implementation Details}

As an official implementation of \RETRO\ is not publicly available, we draw upon the implementation in \citet{norlund-etal-2023-generalization}, which is based on the description in \citet{retro}.
Our implementation deviates only in that (1) we use learnable relative positional biases as in T5 \cite{raffel2020t5}, with a bucket for each unique relative position; (2) instead of BERT \cite{devlin-etal-2019-bert}, we use the pre-trained sentence transformers (ST) \cite{reimers-2019-sentence-bert} model to embed the chunks for the offline retrieval.
ST is preferable over BERT, as it is trained for the task of similarity search, and produces embeddings of lower dimensionality, which makes it more efficient.
We use PyTorch \citep{pytorch} and PyTorch Lightning for distributed training.
For the tokenization, we use the pre-trained T5 tokenizer \cite{t5-hg}.
For retrieving approximate neighbors, we use \texttt{faiss} \citep{faiss}, which performs efficient similarity search between dense representations with GPU support for faster indexing and retrieval.

\subsection{Data}
\label{seq:massiveopentext}

\citet{retro} use the \emph{MassiveText} dataset \cite{gopher} for both training and retrieval.
As this dataset is not publicly available, we set out to replicate it using open sources.
\emph{MassiveText} consists of multilingual text data in five categories: Wikipedia articles, books, GitHub code, news, and common crawl web data.
We use \emph{Pile} \citep{pile} and \emph{RealNews} \citep{realnews} to build a large dataset resembling \emph{MassiveText}'s composition.
The new dataset (see \citet{norlund-etal-2023-generalization} for details) consists of 36M documents containing 52B tokens.
For \emph{Pile}, we keep the training and validation splits, while for \emph{RealNews}, we use the full training set but downsample the validation set to 16,400 news articles to match the proportions of the categories in \emph{Pile}.
For details on the deduplication process, we refer to \citet{pile} and \citet{realnews}.

\subsection{Training}
\label{subsec:training}

We use our dataset to train a \RETRO\ model with approximately 630M parameters. For more details refer to \citet{norlund-etal-2023-generalization}. 
During training, we retrieve from the training set; during validation, we retrieve from the union of the training and validation sets. We train the model on sequences truncated to 1,024 tokens. The chunk size is 64, as in \citet{retro}, and the number of retrieved neighbors is $k=2$ for training and validation. We train the model for 140k training steps with a batch size of 16, taking seven days on 16 A100 GPUs. This means that we use 6\% of the training data during training, not including the retrieved neighbors. As our optimizer, we use Adam \cite{adam} with a fixed learning rate of \num{1e-4}.

\section{A Study on Correlations}

We experiment with two settings: \RETROON, the language model with retrieval enabled, and \RETROOFF, where there are no chunk cross-attention layers and therefore no retrieval, leaving us with a decoder-only language model. As shown by \citet{retro}, the \RETROON\ model performs better when it can exploit an overlap between the generated text and the retrieved neighbor. This is more apparent in text categories with higher token overlap, such as GitHub. The studies in the \RETRO\ paper also show that allowing more overlap when deduplicating the data results in a lower bits-per-byte (\BPB\footnotemark{}). 
\citet{norlund-etal-2023-generalization} take this further to show even minimal overlap results in significant loss reduction, demonstrating the large extent \RETRO\ relies on surface-level similarities.
These findings lead us to hypothesize that having a retrieval method that can find the highest overlapping neighbors will yield lower perplexity (\PPL). Because BERT, ST and similar deep representations of sentences do not always capture surface-level similarities, we set out to investigate where performance gains come from. \footnotetext{\BPB{} $= (L_T/L_B)\mathcal{L}/ln(2)$, where $L_T$ and $L_B$ are the lengths of the validation set in T5 tokens and UTF-8 encoded bytes, respectively, and $\mathcal{L}$ stands for log likelihood loss. $L_T/L_B$ is $0.258415$ for our validation set.}


\begin{table}
    \centering\small
    \begin{tabular}{llcc}
        \toprule
        $X$ & $Y$ & $\rho$ & $r$ \\
        \midrule
        L2$^2$ (ST) & $\Delta \PPL$ & 0.328 & 0.134 \\  
        token overlap & $\Delta \PPL$ & 0.494 & 0.415 \\ 
        \midrule
        L2$^2$ (ST) & token overlap & 0.464 & 0.515 \\ 
        \bottomrule
    \end{tabular}
    \caption{Spearman $\rho$ and Pearson $r$ between variables $X$ and $Y$. L2$^2$ (ST) is the (negative) squared L2 distance between the ST embeddings of $\textsc{Ret}(C_{u-1})$ and $C_{u}$; token overlap is the unigram token overlap between these two chunks; and $\Delta \PPL = \PPL_{\RETROOFF} - \PPL_{\RETROON}$ for the chunk $C_{u}$.}
    \label{tab:corr}
\end{table}

To this end, we measure how the \PPL\ difference ($\Delta \PPL$) between \RETROON\ and \RETROOFF\ for the current chunk ($C_u$, $u \geq 2$) correlates with (1) squared L2 distance between the ST embeddings of $C_u$ and $\textsc{Ret}(C_{u-1})$ (ST), and (2) unigram token overlap, based on T5 tokenization, between $C_u$ and $\textsc{Ret}(C_{u-1})$. The results, reported in Table \ref{tab:corr}, show a considerably stronger correlation between $\Delta \PPL$ and unigram token overlap (measure~2) than between $\Delta \PPL$ and L2 distance (measure~1). The trend is similar between Spearman and Pearson correlation coefficients. 

\section{Changing the Retrieval Method}

As the results from the previous section show a stronger correlation between performance gain and surface-level similarity than ST similarity, we experiment with a retrieval method based on \BMXXV.

\subsection{\BMXXV}
Okapi \BMXXV, introduced by \citet{robertson1995okapi}, is a bag-of-words retrieval method based on tf–idf scores and some free parameters. These parameters are $k_1$, which normalizes the term frequency, and $b$, which controls how much the length of a document would affect the term frequency values. We use Pyserini \cite{Lin_etal_SIGIR2021_Pyserini}, a Python interface to Lucene's \BMXXV\ implementation. We build the \BMXXV{} index on the training set and leave the free parameters at their default values ($k_1 = 0.9$, $b = 0.4$). These values were also shown to perform the best by \citet{karpukhin2020dense}. Using Lucene's Analyzer pipeline\footnote{Lucene Analyzers \cite{lucene-analyzer} are used to extract index terms from text, which includes tokenization and preprocessing.} results in more than 50M unique words for our corpus. We instead use the T5 tokenizer from Hugging Face Transformers \cite{wolf-etal-2020-transformers} and limit our vocabulary to 32k words for the reranking experiments.

\subsection{Retrieving with \BMXXV}

We use the model described in Section~\ref{subsec:training} and change the retrieval method only at inference time to retrieve better neighbors. The results can be found in Table~\ref{tab:ppl}. The perplexity is 14.00 for \RETROOFF\ and 10.87 for \RETROON\ with ST retrieval (\RETROON-ST), corresponding to a 22.3\% reduction in \PPL. Replacing the retrieval method with \BMXXV\ (\RETROON-\BMXXV) gives an additional 13.7\% reduction, which is 61.3\% of the initial drop. For comparability with \citet{retro}, we also report \BPB. 
The results show that using neighbors with more surface-level similarity to the generated chunk is a solid method for leveraging the retrieval mechanism to reduce the perplexity. If the retrieval augmentation is meant to act as an external memory, or to offload memorization from the model \cite{retro}, then \BMXXV\ is a more suitable method to achieve this goal. 



\begin{table}
    \centering\small
    \begin{tabular}{lcc}
        \toprule
        Model & \textsc{Ppl} & \textsc{Bpb} \\
        \midrule
        \RETROOFF & 14.00 & 0.984 \\  
        \midrule
        \RETROON-ST & 10.87 & 0.889 \\  
        \RETROON-ST + \BMXXV\ reranking & 10.46 & 0.875 \\  
        \midrule
        \RETROON-\BMXXV{} & 8.95 & 0.817 \\ 
        \bottomrule
    \end{tabular}
    \caption{\PPL{} and \BPB{} for various retrieval settings on the validation set. The basic \RETRO{} model is the same for all rows.}
    \label{tab:ppl}
\end{table}

\subsection{Reranking}

While the performance gain is significant, finding the \emph{exact} neighbors using \BMXXV{} could be costly, depending on the size of the datasets.
On the other hand, \texttt{faiss} provides an efficient similarity search for dense vectors to find the \emph{approximate} neighbors. Therefore, if enough of the \BMXXV{}-retrieved neighbors could be found among top-$k$ \texttt{faiss}-retrieved ones, with an efficient reranking, we could expect at least part of the performance gain with minimal computational overhead, as long as $k$ is not significantly large. To find an optimal $k$, we first need to know how many of \BMXXV{} neighbors could be found in top-$k$ \texttt{faiss}-retrieved chunks. 

Looking at the \texttt{faiss}-retrieved neighbors, we see that of top-$4$ \BMXXV{}-retrieved neighbors, $17.6\%$ appear in top-$100$ \texttt{faiss}-retrieved chunks, while the overlap is $22.1\%$ for top-$1000$. We decide to continue our experiment with top-$1000$ neighbors, but it is obvious that one could get an even higher overlap with a higher $k$, with diminishing returns. 
The results in Table \ref{tab:ppl} show that with the proposed reranking, \RETROON{}-ST could achieve  $21.3\%$ of the \PPL{} reduction of \RETROON{}-\BMXXV{} compared to \RETROON{}-ST.
The reranking results are interesting not only due to their practical implications but also as an analysis revealing the limited number of high-quality neighbors that can be retrieved using semantic retrieval, even in situations where a large $k$ is feasible.



\section{Related Work}

Augmenting language models with mechanisms that help them incorporate larger contexts has been approached extensively in different forms, such as \citet{guu2018generating}'s retrieve-and-edit approach to reduce the \PPL\ in language generation, and \citet{asai2019learning} that make use of lexical overlap to improve the performance in question answering. 
While retrieval-augmentation has been used with different objectives in mind, such as language modeling \cite{khandelwal20generalization, wu2022memorizing} and machine translation \cite{khandelwal2021nearest}, question answering has been the application to attract the most interest \cite{guu2020retrieval,karpukhin-etal-2020-dense,izacard2021leveraging}. 

An extensive study was performed by 
\citet{izacard2022few}, showing that while we get performance gains using retrieval augmentation, training the retrieval part of the model would yield even more benefits. \RETRO{} \cite{retro}, on the other hand, aims at scaling such language models and therefore opts for keeping the retriever frozen, showing substantial \PPL{} reduction with increasing either the number of language model parameters or the size of retrieval set.

Among the more recent work, \citet{xu2023nearest} found that training using approximate neighbors resulted in a 2.6\% decrease in perplexity. This suggests that non-exact neighbors may have a regularization effect, leading to improved generalization ability. Additionally, \citet{ram2023context} report a drop in perplexity using \BMXXV\ over BERT retrieval using in-context retrieval-augmented language models.

\section{Conclusions and Future Work}

In this paper, we study the source of performance gains in \RETRO, which could be generalized to similar retrieval-augmented language models. 
After observing that the \PPL\ drop correlates more strongly with surface-level overlap between the query and the retrieved text, we replace the retrieval method with \BMXXV, and observe a significant drop in \PPL, which confirms us in the findings of the correlation study. 
This is also an interesting insight as to how these models work, which could be leveraged for performance gain in tasks like question answering where model relies on retrieving facts.
In the end, we also conduct an analysis to find out how much \BMXXV\ neighbors overlap with those retrieved using ST. The results show that while \texttt{faiss} is able to find some of the neighbors with high token overlap, the majority of them remain unretrieved. This is however, enough to gain part of the loss reduction achieved with a pure \BMXXV\ retrieval system.


The proposed methods could also be used during training. By retrieving more overlapping neighbors during training, the process of guiding the model to use retrieved neighbors for language modeling could be done more efficiently. This is particularly relevant when augmenting an already trained language model with a retrieval mechanism. As reported by \citet{retro}, retrieval augmentation results in a larger drop in \BPB\ as the number of model parameters and the size of retrieval data grow. This calls for more efficient methods based on surface-level similarities if we wish to exploit this potential. Furthermore, the retrieval system in \RETRO\ is based on semantic retrieval, the model seems to rely more on surface-level similarities. This could affect the generalizability capabilities of such models, which necessitates further investigations. Lastly, we only evaluate our modified \RETRO\ model on language modeling. It would be interesting to know the impacts of \BMXXV\ retrieval on downstream tasks where retrieval is of use.

\section*{Limitations}
We only experiment with one type of retrieval-augmented language models, i.e., \RETRO. However, the ways the other models retrieve neighbors and integrate them are not so much different to affect the results in this paper. The experiments in this paper are done with a small size \RETRO\ model and data compared to the sizes considered by \citet{retro}, due to computational limitations. According to the same authors, however, the gains should be constant with the increase of the model and retrieval set size. The larger models are mainly different in their behavior when there is no overlap. However, this should not affect the \emph{copying} tendency of these models tremendously, as it is still the easiest way to generate the next token. It is also worth noting that \RETROOFF, while not using retrieval at test time, is still \emph{trained} using retrieval -- so it is not a complete retrieval-free model. The results presented by \citet{retro} however, show that \RETROOFF\ is on a par with their retrieval-free baseline in terms of \BPB. Finally, we note that our evaluations have only considered the perplexity under teacher forcing, and we have not investigated the behavior of the model in free-form generation or with any kind of fine-tuning.

\section*{Acknowledgements}
This work was partially supported by the Wallenberg AI, Autonomous Systems and Software Program (WASP) funded by the Knut and Alice Wallenberg Foundation.
The computations were enabled by resources provided by the National Academic Infrastructure for Supercomputing in Sweden (NAISS) at Alvis partially funded by the Swedish Research Council through grant agreement no.\ 2022-06725, and by the Berzelius resources provided by the Knut and Alice Wallenberg Foundation at the National Supercomputer Center.

\bibliography{anthology,custom}
\bibliographystyle{acl_natbib}

\end{document}